%% file: main.tex
\documentclass[letterpaper, 10 pt, conference]{ieeeconf}

\IEEEoverridecommandlockouts

\usepackage{etextools}
\usepackage{amssymb}
\usepackage{float}
\usepackage{tabto}
\usepackage{makeidx}
\usepackage{amsmath}
\usepackage{bbm}
\usepackage{epsfig}
\usepackage{epsf}
\usepackage{psfrag}
\usepackage{verbatim}
\usepackage{color}
\usepackage{multirow}
\usepackage{tabularx}
\usepackage{booktabs}
\usepackage[tight,footnotesize]{subfigure}
\usepackage{array}
\usepackage{soul}
\usepackage{footnote}
\usepackage{cite}
\usepackage{dblfloatfix}
\usepackage{changepage}

\usepackage{color, colortbl}
\usepackage[colorlinks,bookmarksnumbered,citecolor=orange,urlcolor=orange]{hyperref}
\usepackage{graphicx}
\graphicspath{{Figures}}
\DeclareGraphicsExtensions{.pdf,.png}
\usepackage{bigstrut}
\usepackage[english]{babel}

\usepackage{csquotes}

\usepackage[T1]{fontenc}
\usepackage{algorithm, setspace}
\usepackage{algpseudocode}
\usepackage{url}
\usepackage{multirow}
\usepackage{float}
\usepackage{xcolor}
\usepackage{hyperref}
 \hypersetup{
     colorlinks=true,
     linkcolor=orange,
     filecolor=orange,
     citecolor=orange,      
     urlcolor=orange,
     }

\newcommand{\p}[1]{\smallskip \noindent \textbf{{#1}.}}

\newcommand{\fig}[1]{Figure~\ref{fig:#1}}
\usepackage{balance}

\title{\LARGE

Kiri-Spoon: A Soft Shape-Changing Utensil for Robot-Assisted Feeding}

\author{Maya N. Keely, Heramb Nemlekar, and Dylan P. Losey
\thanks{This work is supported in part by NSF Grant $\#2205241$.}
\thanks{The authors are members of the Collaborative Robotics Lab (\href{https://collab.me.vt.edu/}{Collab}), Dept. of Mechanical Engineering, Virginia Tech, Blacksburg, VA 24061.
\newline
{e-mail: \texttt{\{mayakeely, hnemlekar, losey\}@vt.edu}}}
}

\begin{document}
\maketitle

\begin{abstract}

Assistive robot arms have the potential to help disabled or elderly adults eat everyday meals without relying on a caregiver.
To provide meaningful assistance, these robots must reach for food items, pick them up, and then carry them to the human's mouth.
Current work equips robot arms with standard utensils (e.g., forks and spoons).
But --- although these utensils are intuitive for humans --- they are not easy for robots to control.
If the robot arm does not carefully and precisely orchestrate its motion, food items may fall out of a spoon or slide off of the fork.
Accordingly, in this paper we design, model, and test \textit{Kiri-Spoon}, a novel utensil specifically intended for robot-assisted feeding.
Kiri-Spoon combines the familiar shape of traditional utensils with the capabilities of soft grippers.
By actuating a kirigami structure the robot can rapidly adjust the curvature of Kiri-Spoon: at one extreme the utensil wraps around food items to make them easier for the robot to pick up and carry, and at the other extreme the utensil returns to a typical spoon shape so that human users can easily take a bite of food.
Our studies with able-bodied human operators suggest that robot arms equipped with Kiri-Spoon carry foods more robustly than when leveraging traditional utensils.
See videos here: https://youtu.be/nddAniZLFPk

\end{abstract}


\input{1_intro}
\input{2_related}
\input{3_kirispoon}

\input{4_models}
\input{5_userstudy}

\input{6_conclusion}


\balance
\bibliographystyle{IEEEtran}
\bibliography{references}

\end{document}

%% file: 1_intro.tex
\section{Introduction} \label{sec:intro}

\begin{figure}[ht!!]
    \begin{center}
        \includegraphics[width=0.9\columnwidth]{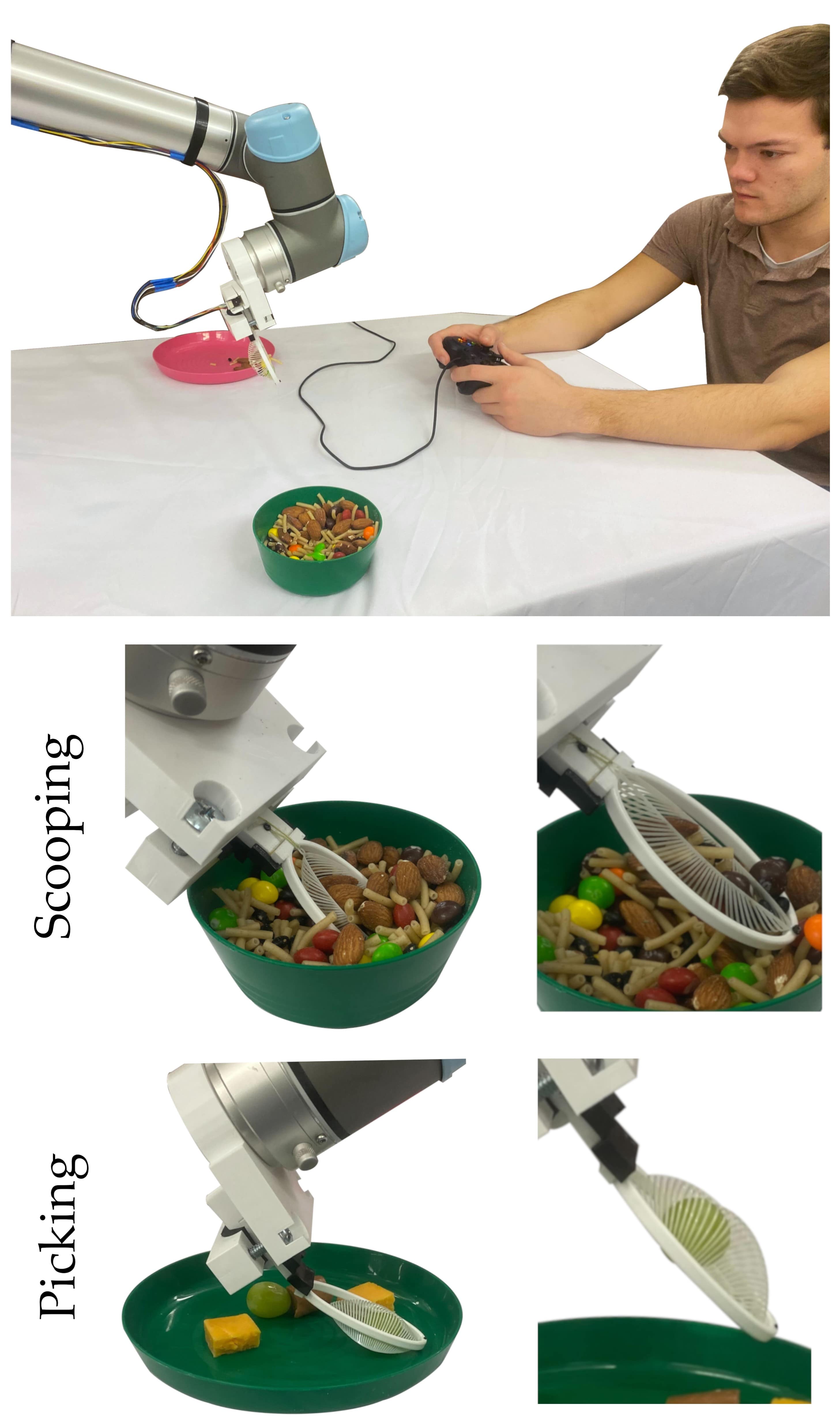}
        \caption{Human controlling the robot arm and Kiri-Spoon during robot-assisted feeding. The operator relies on the robot arm to pick up and carry food items. Our proposed Kiri-Spoon makes this easier by encapsulating and releasing foods within a soft kirigami structure with adjustable curvature. Using Kiri-Spoon the robot can scoop food from a bowl or pick food off a plate while wrapping around that food firmly to prevent spills.}
        \label{fig:front}
    \end{center}
    \vspace{-2.0em}
\end{figure}

Over $1.8$ million American adults living with motor impairments need assistance when eating \cite{taylor2018americans}.
Assistive robots --- such as wheelchair-mounted robot arms \cite{argall2018autonomy} --- have the potential to help these users eat their everyday meals and regain independence \cite{nanavati2023design}.
Consider the robot-assisted feeding scenario shown in \fig{front}.
To seamlessly provide assistance, the robot arm must be able to pick up different food items and then carry them to the human's mouth.
Humans typically use utensils (e.g., forks or spoons) to pick up and transfer food.
But as we develop robot arms to help automate feeding tasks, \textit{what utensils should assistive robots use?}

Existing research answers this question by focusing on either the human's preferences or the robot's capabilities.
From the \textit{human's perspective}, traditional utensils like forks or spoons are convenient for transferring food from the robot to the human (e.g., taking a bite from a fork).
Works on robot-assisted feeding \cite{gordon2023towards, park2020active, losey2022learning, ondras2023human, schultz2022proof, feng2019robot, belkhale2022balancing} have therefore equipped the robot with traditional utensils, and then explored how robot arms should control those utensils.
However, it still remains challenging for robots to carefully and precisely manipulate human-friendly utensils: morsels may slide off forks or fall out of spoons.
From the \textit{robot's perspective} we can therefore make food handling more robust by introducing new end-effector designs.
Recent soft grippers such as \cite{shintake2018soft, wang2022challenges, mehta2023riso, gafer2020quad, zhu2023bioinspired, ruotolo2021grasping} enhance robot capabilities by encapsulating, holding, or adhering to diverse sets of food items in ways that forks or spoons cannot achieve.
But this increase in robot capability comes at the cost of human convenience: today's soft grippers are not utensils, and users cannot easily transfer foods from these end-effectors to their mouths.

In this paper, we seek to unite human and robot perspectives by developing a novel type of utensil specifically for robot-assisted feeding.
In order to balance the human's needs with the robot's capabilities, our hypothesis is that:
\begin{center}\vspace{-0.3em}
\textit{Utensils for assistive robots should be similar to traditional utensils in \emph{shape}, and similar to soft grippers in \emph{function}}.
\vspace{-0.3em}
\end{center}
Based on this hypothesis we introduce Kiri-Spoon: a spoon-shaped utensil that leverages a soft kirigami gripping mechanism (see \fig{front}).
In practice the assistive robot arm first positions and rotates Kiri-Spoon as it would a traditional spoon (scooping under the food morsel) or a traditional fork (aligning above the food morsel).
The robot then actuates the kirigami structure to rapidly increase the curvature of Kiri-Spoon, creating a soft bowl that wraps around and pinches the desired food.
In this high-curvature state Kiri-Spoon is more robust than standard utensils --- i.e., items do not slide off or fall out of the kirigami structure.
Finally, the robot arm uses Kiri-Spoon to carry the morsel to the human; here we can decrease the kirigami curvature and revert to a typical spoon shape for human-friendly assistive feeding.  

Overall, this paper presents our first steps towards an assistive feeding utensil that balances the human's preferred shape with advances in soft robotic gripping. We make the following contributions:

\p{Designing Kiri-Spoon} 
We present the design process behind Kiri-Spoon. 
The key component of Kiri-Spoon is a soft, elliptical sheet of polymer resin with parallel cuts.
When this kirigami structure is pulled on both ends it deforms into a bowl of increasing curvature.

\p{Modeling the Mechanics and Geometry}
We model Kiri-Spoon as a spring-loaded four-bar linkage, where the spring stiffnesses depend on
the design parameters.
Our experiments show that this model can accurately predict the force needed to deform the kirigami structure to a desired curvature.

\p{Comparing to Traditional Utensils}
We conduct an initial user study where $12$ able-bodied participants control a robot arm to pick up food items from plates and bowls. 
When the robot uses forks or spoons, these food items often spill onto the table.
Kiri-Spoon reduces these spills, resulting in a system that is subjectively preferred by human users.

%% file: 2_related.tex
\begin{figure*}[t]
    \begin{center}
        \includegraphics[width=1\textwidth]{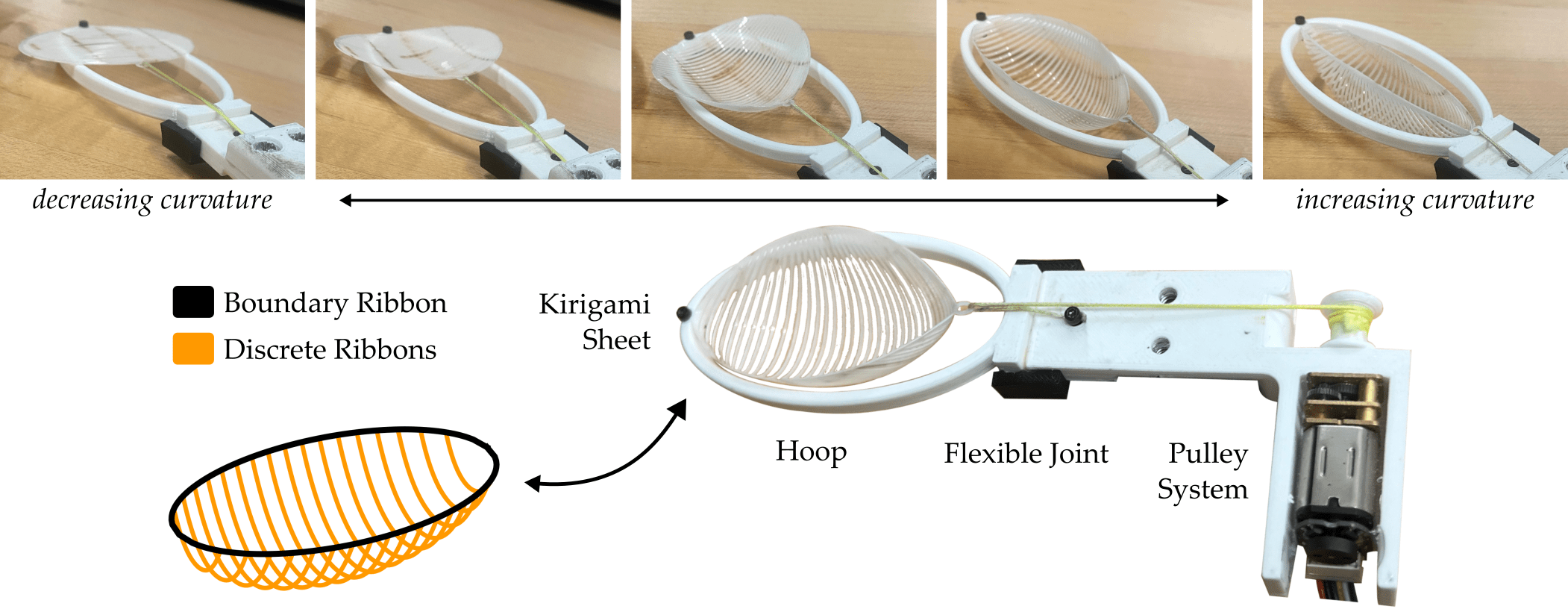}
        \caption{Kiri-Spoon is composed of a $2$D kirigami sheet actuated by a pulley. (Left) The food-safe sheet is cut into an ellipse with one boundary ribbon and multiple discrete ribbons.
        (Top) When forces are applied to the ends of the boundary ribbon, the discrete ribbons buckle and the $2$D sheet morphs into a $3$D bowl with adjustable curvature. (Right) The flexible joint enables the entire Kiri-Spoon to bend when colliding with objects (e.g., a plate or bowl).}
        \label{fig:spoon}
        \vspace{-3 ex}
    \end{center}
\end{figure*}

\section{Related Work} \label{sec:related}

\noindent \textbf{Robot-Assisted Feeding.} 
Our work is intended for scenarios where a person with motor impairments is using an assistive robot arm to eat everyday meals.
There are several challenging aspects of this setting, including determining how users can intuitively control assistive robot arms \cite{rea2022still}, 
identifying how robots should blend user commands and autonomous actions \cite{bhattacharjee2020more}, 
and learning how the robot arm can help reach for and grasp food items \cite{feng2019robot}.
In this paper we specifically focus on the problem of picking up and carrying food morsels for robot-assisted feeding.
Here recent research suggests that users prefer traditional utensils like forks and spoons, particularly in social contexts \cite{nanavati2023design}.
However, assistive robot arms struggle to manipulate these traditional utensils with the same dexterity and effectiveness as able-bodied human operators \cite{losey2022learning, gordon2023towards}.
Related works therefore tackle this trade-off by either (a) improving the algorithm the robot uses to control traditional utensils, or (b) developing novel end-effectors to improve the mechanics of food handling.

\p{Using Traditional Utensils}
We first survey works that equip the robot with a fork or spoon, and then develop learning and control strategies to manipulate that utensil. 
To start the process the robot uses a camera and user interface to detect which food item the human wants \cite{park2020active, schultz2022proof}.
The robot then fully or partially automates the motion of reaching for and picking up that food morsel.
If the robot is autonomous, methods such as \cite{gordon2023towards, feng2019robot} learn to orient the robot's utensil to skewer the desired food. 
Alternatively --- if the system is partially automated --- approaches like \cite{losey2022learning} map the operator's inputs to scooping or cutting motions.
Once the robot has grasped the food with its utensil, it then plans a trajectory to carry that food to the human's mouth \cite{belkhale2022balancing}.
Finally, the robot arm carefully positions and orients the utensil so the human can easily and safely take a bite from the utensil \cite{ondras2023human}.
Two points of failure in this process are acquisition (i.e., \textit{can the robot robustly pick up diverse foods?}) and transfer (i.e., \textit{can the robot carry those foods without them falling or spilling?}).
Our work proposes an alternative utensil that could be used with the same algorithms described above, while making it easier for robots to grasp and carry foods.

\p{Soft Grippers for Food Handling}
State-of-the-art research on soft grippers has introduced a variety of designs to meet the needs of the food processing industry \cite{shintake2018soft, wang2022challenges}.
For example, \cite{gafer2020quad} propose a robotic hand with four compliant fingers that slide under the food and enclose it within a box. 
Alternatively, \cite{mehta2023riso} control switchable and soft adhesives to make foods stick to the bottom of a rigid gripper.
In general, these soft finger-like mechanisms \cite{zhu2023bioinspired, ruotolo2021grasping} outperform traditional utensils in terms of picking up, holding, and handling diverse food items.
But state-of-the-art soft grippers are not designed for assistive eating; it is not easy to transfer foods from these grippers to the human's mouth.
Returning to our examples, with either \cite{gafer2020quad} or \cite{mehta2023riso} the robot would need to drop food items from the soft gripper into the human's mouth, or the robot would need a secondary device specifically for bite transfer.
Most related to our approach are works on robot-assisted feeding that \textit{slightly} modify a traditional utensil.
In \cite{sundaresan2023learning} and \cite{shaikewitz2023mouth} the authors add degrees-of-freedom to a fork, and in \cite{grannen2022learning} the robot uses two arms: one arm with a spoon, and another to help push foods onto that spoon.
Like these approaches we plan to modify the robot's utensil.
However, we will apply recent advances in soft gripper mechanics to develop a new utensil type that is like a spoon in form, but like a soft gripper in function.

%% file: 3_kirispoon.tex
\section{Kiri-Spoon Design} \label{sec:design}

While traditional feeding utensils like spoons and forks are intuitive for humans to use, they often lack the ability to securely hold the acquired food. On the other hand, soft grippers can encapsulate and firmly adhere to foods but are not well-suited for feeding humans.
Accordingly, we here introduce a novel feeding utensil (\textit{Kiri-Spoon}) which serves both functions: (i) maintaining a familiar spoon-like shape when collecting the food and (ii) encapsulating the acquired food to avoid spills during transfer. 

In this section we introduce the components of Kiri-Spoon and explain how it is actuated to collect and grasp food items. We also detail the process for fabricating Kiri-Spoon and discuss key design considerations for the kirigami structure.

\p{Components}
Figure~\ref{fig:spoon} showcases our proposed Kiri-Spoon design.
The main component of Kiri-Spoon is a shape-morphing \textit{kirigami sheet}~\cite{hong2022boundary} which forms the bowl of the spoon. 
This kirigami structure is a 2D sheet with parallel cuts that create multiple discrete ribbons surrounded by one boundary ribbon. When the boundary is pulled orthogonal to the discrete ribbons the 2D sheet morphs into a 3D structure due to out-of-plane buckling of the discrete ribbons. 
The shape of the resulting 3D structure depends on the initial boundary ribbon. 
We will use an elliptical kirigami sheet to achieve a spoon-like shape when deformed.

One end of the kirigami sheet is secured to a \textit{rigid hoop} while the other end is tied to a string controlled by a motor-driven \textit{pulley system}. Specifically, the sheet is oriented such that the discrete ribbons are perpendicular to the direction in which the string is pulled.
The spoon handle is connected to the rigid hoop through a \textit{flexible joint} that allows the hoop to bend when pressed against the food or the container. This feature protects the hoop from breaking and helps to align the kirigami sheet over the food item.

\p{Actuation} 
When the motor is actuated, the string pulls one end of the elliptical boundary, increasing its length in the direction of the applied force and decreasing its width along the discrete ribbons. This compresses the discrete ribbons, causing them to buckle and create a spoon-like shape.

In this configuration, the Kiri-Spoon can scoop food from a bowl, similar to a traditional spoon. Upon collecting the food, the boundary can be pulled further to enclose or grasp the food by reducing the boundary width. This ability to deform and grasp food items allows Kiri-Spoon to also function as a utensil for picking up food from flat surfaces, offering an alternative to using a fork, as shown in Figure~\ref{fig:front}.

\begin{figure*}[ht]
    \begin{center}
        \includegraphics[width=\textwidth]{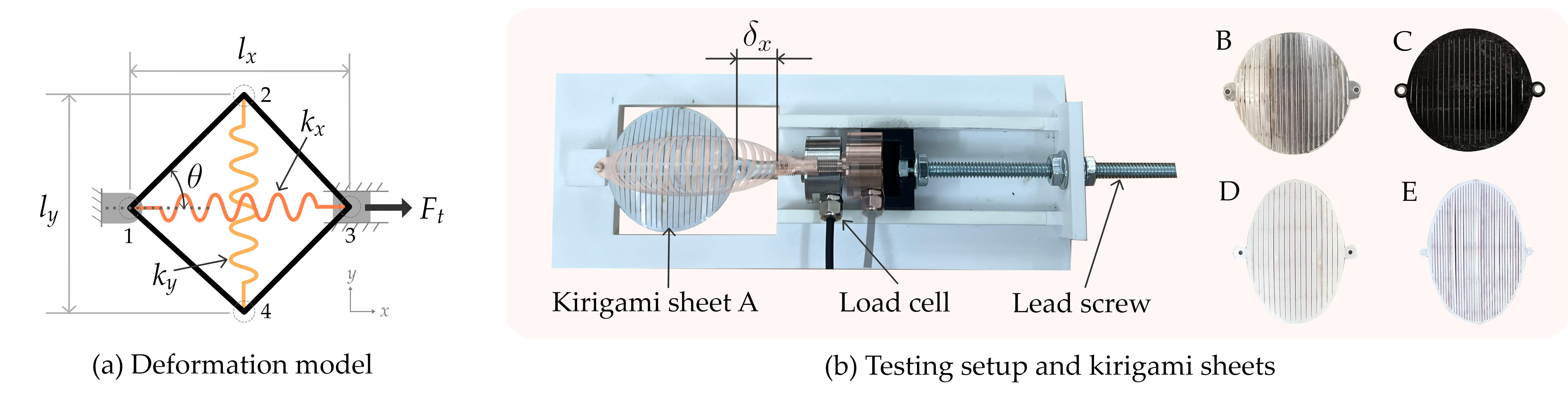}
        \vspace{-4 ex}
        \caption{(a) Spring-loaded four-bar linkage. The dimensions $l_{x}$ and $l_{y}$ represent the principal axes of the elliptical kirigami sheet. The tensile force $F_{t}$ displaces the slider, increasing the curvature of the kirigami structure. This displacement is opposed by two springs with constants, $k_{x}$ and $k_{y}$, that model the bending stiffness of the boundary and the reaction force of the discrete ribbons, respectively. (b) Testing setup for validating our proposed deformation model. The kirigami sheets A, B, and C have a circular boundary with a radius of $23.5$ mm, while sheets D and E have an elliptical boundary with the same dimensions described in Section~\ref{sec:design}. All sheets are made from PET, except sheet C which is made from thermoplastic polyurethane (TPU). The thickness of each sheet and the width of their discrete ribbons are specified in Table~\ref{tab:stiffness}.}
        \label{fig:modeling}
        \vspace{-3 ex}
    \end{center}
\end{figure*}

\p{Material}
The material of the kirigami sheet impacts the rigidity of the structure and the force required to deform it. For our application, the material needs to be \textit{isotropic} so that the sheet deforms uniformly when pulled. The material should also have sufficient \textit{ductility} to deform substantially without breaking, and also be \textit{elastic} enough to return to the original shape when released. We therefore fabricate the kirigami sheets for Kiri-Spoon with polyethylene terephthalate (PET), an inexpensive and \textit{food-safe} plastic that exhibits the desired material properties.

\p{Fabrication}
We used laser cutting to shape a $0.25$ mm thick PET sheet into an ellipse with a minor axis of $17.8$ mm and a major axis of $26.7$ mm.
Using the same laser cutting process, we created discrete ribbons of $1$ mm width by making slits parallel to the ellipse's major axis.
We selected these dimensions such that, when the kirigami sheet is deformed, the shape and volume of the resulting 3D structure are comparable to that of a conventional dessert spoon. In particular, we designed an ellipse with an aspect ratio of $1.5$, with the discrete ribbons parallel to the major axis and perpendicular to the minor axis along which the boundary is pulled. This configuration allows the discrete ribbons to have a greater range of deformation and enables Kiri-Spoon to grasp wider food items. 
The tensile force required to deform the kirigami sheet increases with increasing thickness of the sheet and the width of the discrete ribbons. Accordingly, we chose a thin PET sheet and created discrete ribbons of just $1$ mm thickness to minimize the force required by the motor-driven system to actuate the kirigami structure.

In the following sections, we will model and validate the deformation of the kirigami structure, and then compare our proposed Kiri-Spoon design to traditional utensils in a robot-assisted feeding task.

%% file: 4_models.tex
\section{Deformation Model} \label{sec:models}

Kiri-Spoon starts as a 2D elliptical kirigami sheet that morphs into a 3D spoon-shaped structure when a tensile force is applied to its boundary, orthogonal to the discrete ribbons.
The change in the curvature of the kirigami structure is governed by complex structural deformations that include the bending and torsion of the boundary ribbon and the bucking of the discrete ribbons~\cite{hong2022boundary}. 
In this section, we propose a simplified model to approximate the deformation of kirigami sheets corresponding to the applied tensile force. We first present how we predict the geometry of the boundary and discrete ribbons, and then outline our approach for modeling the tensile force needed to deform the kirigami structure.

\p{Deformation of Boundary Ribbon}
We model each kirigami sheet as a spring-loaded four-bar mechanism shown in Figure~\ref{fig:modeling}-(a). 
The dimensions of our model, $l_x$ and $l_y$, correspond to the lengths of the principal axes of the elliptical boundary of the kirigami sheet. Here, $l_x$ represents the length of the sheet perpendicular to the discrete ribbons and $l_y$ represents the width of the sheet parallel to the discrete ribbons.

To simulate the deformation of a kirigami sheet, we first measure the lengths of the principal axes of its elliptical boundary in the undeformed state. We then set these lengths as the initial dimensions, $l_{x, init}$ and $l_{y, init}$, of the four-bar linkage in the absence of tensile force and compute the length of each link.
\begin{align}
    l_{link} &= l_{1-2} = l_{2-3} = l_{4-3} = l_{1-4} \nonumber\\
    &= \sqrt{(l_{x, init} / 2)^{2} + (l_{y, init} / 2)^{2}}
\end{align}

Joint $1$ is anchored to a rigid support. When a tensile force $F_{t}$ is applied to the slider at joint $3$, it moves away from joint $1$, increasing the length of the structure along the x-axis. Consequently, the rigid links pull joints $2$ and $4$ closer to each other, decreasing the width of the structure along the y-axis.
Because the link lengths remain constant, we can determine the new dimensions of the four-bar linkage when joint $3$ is displaced by any distance $\delta_x$.
\begin{equation}
    l_x = l_{x, init} + \delta_{x} \label{eq:length}
\end{equation}
\begin{equation}
    l_y = \sqrt{l_{link}^{2} - (l_{x}/2)^{2}} \label{eq:width}
\end{equation}

We propose that these lengths correspond to the actual lengths of the principal axes of the elliptical boundary after deformation. Assuming the origin is at joint $1$, we approximate the boundary ribbon's shape with an ellipse centered at $(0, (l_{x}/2))$, where $(x, y)$ are the boundary coordinates.
\begin{equation}
    \frac{(x - (l_{x}/2))^2}{(l_{x}/2)^{2}} + \frac{y^2}{(l_{y}/2)^{2}} = 1 \label{eq:ellipse}
\end{equation}

\begin{figure*}[ht]
    \begin{center}
        \includegraphics[width=1\textwidth]{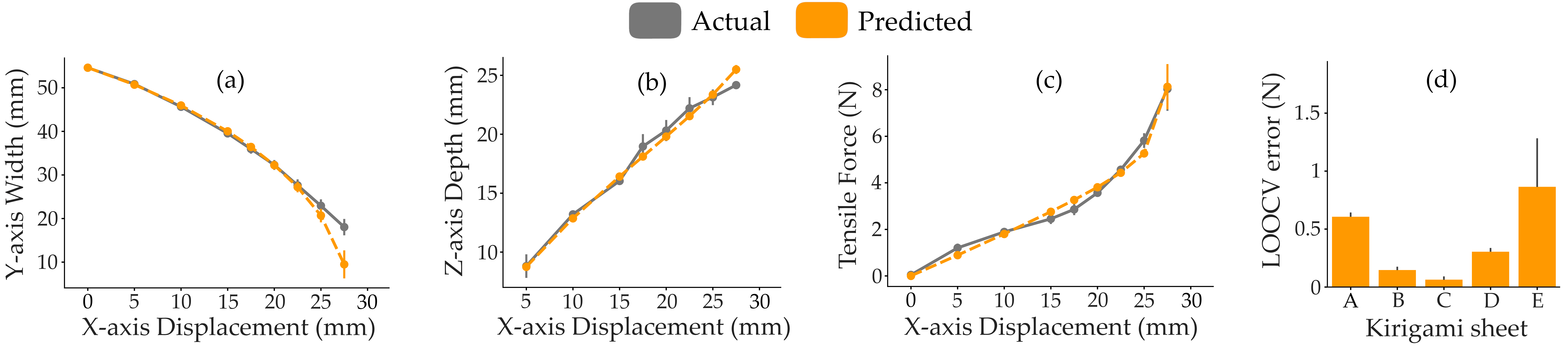}
        \vspace{-4 ex}
        \caption{Model validation results. (a)-(c) The predicted width and depth of the kirigami sheet E and the corresponding tensile force for each displacement along the x-axis. These results show that our model predictions closely approximate the actual measurements of the dimensions and tensile forces for sheet E. (d) Leave-one-out cross-validation (LOOCV) results for each sheet. The mean absolute error in the estimated tensile force was at most $0.86$ N.}
        \label{fig:validate}
        \vspace{-3 ex}
    \end{center}
\end{figure*}

\p{Deformation of Discrete Ribbons}
At the start, the discrete ribbons lie in the same plane as the boundary ribbon. When the boundary is deformed, reducing its width along the y-axis, the endpoints of each discrete ribbon move closer, bending the ribbon to form a downward arch. We will approximate this shape as a catenary and obtain the following relationship between the length of a discrete ribbon $l_{ribbon}$ and the distance between its endpoints $d_{y}$.
\begin{equation}
    l_{ribbon} = 2a\sinh(d_{y} / 2a) \label{eq:catenary_length}
\end{equation}

The parameter $a$ characterizes the shape of the catenary. To determine this shape, we first identify the endpoints of the discrete ribbon by substituting its x-coordinate and the boundary dimensions in Equation~\ref{eq:ellipse} and computing the roots. We calculate the distance between these roots to obtain $d_{y}$. Then, we numerically solve Equation~\ref{eq:catenary_length} to obtain the value of $a$ and model the shape of the discrete ribbon.
\begin{equation}
    z = a\cosh(y/a) - (a + d_{z}) \label{eq:catenary}
\end{equation}
Here, $(y, z)$ are the discrete ribbon coordinates and the term $(a + d_{z})$ is an offset that ensures that its endpoints remain in the same plane as the boundary, i.e., $z=0$. The distance $d_z$ is the maximum depth of the discrete ribbon along the z-axis and can be obtained from the following property of catenaries with both endpoints at the same height.
\begin{equation}
    d_{z}^{2} + 2ad_{z} - (l_{ribbon}/2)^{2} = 0 \label{eq:catenary_depth}
\end{equation}

The depth of each discrete ribbon is directly proportional to its length. The longest ribbon is at the center of the ellipse and determines the maximum depth of the kirigami structure along the z-axis, denoted as $l_{z}$. The series of catenaries formed by subsequent ribbons of decreasing lengths result in the desired spoon-shaped structure shown in Figure~\ref{fig:spoon}.

\p{Applied Tensile Force}
When a tensile force is applied for bending the boundary ribbon, the buckled discrete ribbons exert a reaction force on the boundary, aiming to restore it back to its original width. As a result, the applied tensile force must overcome both the bending stiffness of the boundary and the reaction from the discrete ribbons.

We model these opposing forces with two springs, $S_{1-3}$ and $S_{2-4}$, along the principal axes of the elliptical boundary. The spring connected to joints $1$ and $3$ represents the bending stiffness of the boundary ribbon, while the spring connected to joints $2$ and $4$ models the equivalent reaction force of the discrete ribbons. 
When the slider is displaced by $\delta_x$ from the initial resting position, the spring $S_{1-3}$ exerts a proportional force $F_{x} = k_{x}\delta_{x}$ in the opposite direction. The movement of the slider compresses the spring $S_{2-4}$ which exerts a reaction force of $F_{y} = k_{y}\delta_{y}$ along the y-axis, where $\delta_{y} = l_{y, init} - l_{y}$. The rigid links connected to $S_{2-4}$ transmit a component of this reaction force to the slider.
\begin{equation}
	F_{1-2} + F_{2-3} = F_{1-4} + F_{4-3} = \frac{k_{y}\delta_{y}}{\sin\theta}
\end{equation}

As our model is symmetric, the force $F$ in each rigid link is equal.  The force transmitted through links $2-3$ and $4-3$ exerts an opposing force on the slider at an angle $\theta$, which is the angle between the links and the applied tensile force. Thus, the tensile force required to displace the slider by $\delta_{x}$ is equal to the sum of all the opposing forces.
\begin{equation}
    F_{t} = k_{x}\delta_{x} + (k_{y}\delta_{y}/ \tan\theta) \label{eq:tensile_force}
\end{equation}

Here, $\theta = \arctan(l_{y} / l_{x})$. The spring constants $k_{x}$ and $k_{y}$ depend on the material and thickness of the kirigami sheet as well as the width of the discrete ribbons. In the following section, we validate the proposed deformation model and describe our approach for determining the spring constants for different kirigami sheets.

\section{Model Validation}\label{sec:validate}

Our proposed model approximates the complex dynamics governing the deformation of the kirigami structure when subjected to a tensile force. We now evaluate whether the dimensions and forces predicted by our model are consistent with the deformation of kirigami sheets in practice.

We tested with five kirigami sheets, each having different dimensions and material properties as shown in Figure~\ref{fig:modeling}-(b). Our testing setup consists of a load cell connected to a lead screw. Each kirigami sheet is attached to the fixed support on one side and to the load cell on the other. Starting from a position where the sheet is undeformed, we displaced the lead screw in increments of $2.5-5$ mm along the x-axis. For each displacement $\delta_{x}$, we recorded the length $l_{x}$, width $l_{y}$, and depth $l_{z}$ of the deformed sheet, and the corresponding tensile force $F_t$ measured by the load cell.


\p{Deformation of Boundary Ribbon} 
We first compare the actual boundary widths for each displacement to the widths estimated by our model ($l_y$) based on initial lengths $l_{x, init}$ and $l_{y, init}$ of the kirigami sheets. Figure~\ref{fig:validate}-(a) shows the actual and predicted widths of the boundary for sheet $E$, which is the same kirigami sheet that we used to design Kiri-Spoon in Section~\ref{sec:design}. As the displacement increases, both the actual and predicted widths decrease in a similar manner. The model predictions deviate from the actual measurements only for large displacements. This deviation occurs because, at high strains, the boundary ribbon not only bends but also stretches, leading to an increased perimeter and consequently a greater width than what our model predicted. Despite this, the mean absolute error in estimating the boundary width across all sheets and displacements ($\geq 5$ mm) was only $1.78$~mm. Additionally, the coefficient of determination was $R^{2} = 0.915$, indicating that our proposed four-bar linkage model was able to explain $91.5\%$ of the variability in the width of the boundary ribbon. 

\p{Deformation of Discrete Ribbons} We now compare the depths predicted by our model ($l_{z}$) to the actual depths of the kirigami structures for each displacement. This is the maximum depth $d_{z}$ of the discrete ribbon at the center of each kirigami sheet. Figure~\ref{fig:validate}-(b) shows that the predicted depths closely align with the actual depths measured for sheet $E$. Across all sheets, the mean absolute error in predicting the depths was only $0.84$ mm with a coefficient of determination $R^{2} = 0.95$. This indicates that our approach of approximating the discrete ribbons as catenaries explained $95\%$ of the variability in the maximum depths of the kirigami structures along the z-axis. 

Overall, these results suggest that our deformation models closely approximate the dimensions of kirigami structures. 

\begin{table}[t]
\caption{Modeled spring stiffness (N/m)}
\label{tab:stiffness}
\vspace{-1 ex}
\resizebox{\columnwidth}{!}{%
\begin{tabular}{c|c|c|c|c|c}
\hline
Sheet & Material & Thickness & Ribbon width & $\mathbf{k_x}$ & $\mathbf{k_y}$ \\ \hline
A & PET & 0.25 mm & 2 mm & 320.02 & 54.07 \\
B & PET & 0.15 mm & 2 mm & 76.15 & 28.74 \\
C & TPU & 1 mm & 2 mm & 76 & 3.95 \\
D & PET & 0.25 mm & 2 mm & 184 & 11.17 \\
E & PET & 0.25 mm & 1 mm & 171.78 & 9.25 \\ \hline
\end{tabular}
}
\vspace{-1 ex}
\end{table}

\p{Applied Tensile Force}
Lastly, we assess whether our proposed model can predict the tensile forces required to deform the kirigami sheets. For each sheet, we first computed the dimensions of the four-bar linkage ($l_x$ and $l_y$), the displacement along the y-axis ($\delta_{y}$), and the angle ($\theta$) corresponding to each x-axis displacement, following our approach in Section~\ref{sec:models}. We then substituted these values along with the actual tensile forces in Equation~\ref{eq:tensile_force} to obtain a set of linear equations in the spring constants $k_x$ and $k_y$. We solved these equations using the least squares method to compute the spring constants and model the tensile forces for that sheet.

Figure~\ref{fig:validate}-(c) shows the actual and predicted tensile forces for sheet E. The predicted tensile force increases linearly for small displacements and exponentially for larger displacements, similar to the measured force. Table~\ref{tab:stiffness} presents the spring constants computed for each kirigami sheet. As discussed in Section~\ref{sec:design}, we expect the required tensile force to increase with the thickness of the kirigami sheet and the width of discrete ribbons. Accordingly, we observe that the spring constants for the $1$ mm thick sheet B are lower than those for the $2$ mm thick sheet A. 

The spring constants for sheet C are even lower than sheets A and B due to the less rigid nature of TPU compared to PET. While this reduces the force required to deform the sheet, it also compromises its ability to maintain a rigid spoon-like shape when deformed. When comparing sheets D and E, we notice a decrease in the spring constants as the width of the discrete ribbons reduces from $2$ mm to $1$ mm. This occurs because thinner discrete ribbons exert a smaller reaction force on the boundary, making it easier to deform the sheet while maintaining its rigidity due to the properties of the PET material.

To evaluate our model, we performed leave-one-out cross-validation (LOOCV) for each kirigami sheet. This involved predicting the tensile force for each displacement while excluding it from the data when solving for the spring constants and then comparing the predicted force to the actual force measured for that displacement. Figure~\ref{fig:validate}-(d) shows the mean absolute error in the predicted tensile forces for each kirigami sheet. Overall, the mean error in predicting the tensile force was at most $0.86$ N, indicating that our proposed spring-loaded model effectively approximated the tensile forces required to deform the kirigami sheets.

\textit{Design Implications.} We used our model predictions for deformation of the boundary and discrete ribbons to determine the size of food items that each structure can hold and accordingly decided the boundary dimensions of the kirigami sheet in our Kiri-Spoon design. Additionally, we used the tensile force predictions to decide the thickness of the kirigami sheet and width of the discrete ribbons when designing Kiri-Spoon, ensuring that our motor-driven pulley system can generate the required force.

%% file: 5_userstudy.tex

\section{User Study} \label{sec:userstudy}

\begin{figure*}[t]
    \begin{center}
        \includegraphics[width=2.0\columnwidth]{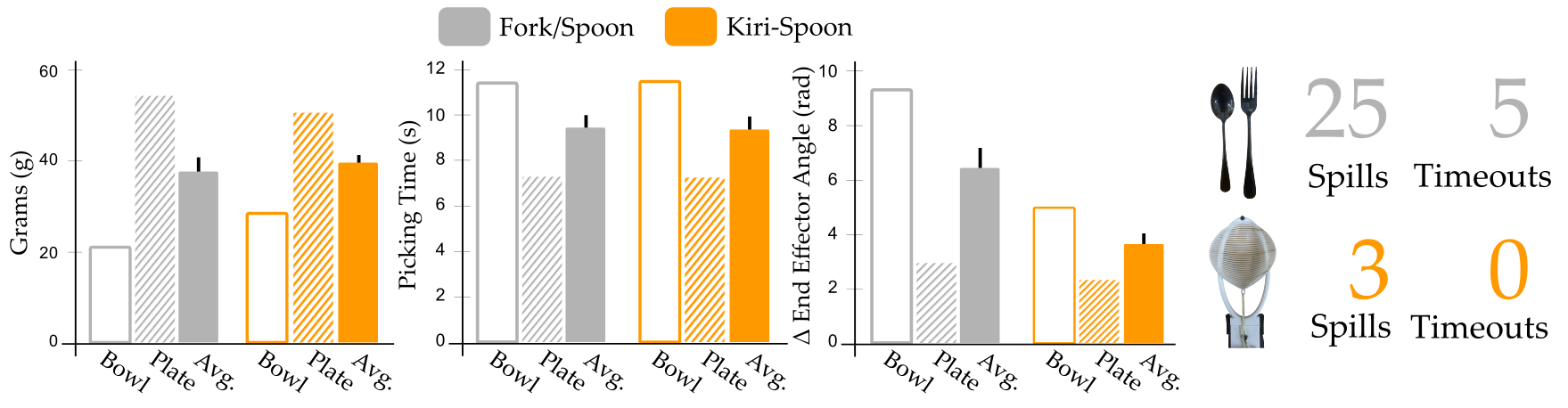}
        \vspace{-1 ex}
        \caption{Objective results from the user study. In this study participants controlled a robot arm to scoop food from a bowl and pick food from a plate using feeding utensils attached to the robot's end-effector. Participants then controlled the robot to carry the food across the table and finally drop the food in another container. We compared traditional utensils (e.g., forks and spoons) to our Kiri-Spoon. We found that while users transferred similar amounts of food (\textit{Transferred Weight}) and spent a similar amount of time picking the food (\textit{Picking Time}) with both types of feeding utensils, they spent significantly less time orienting the end effector (\textit{Orient}), had no \textit{Timeouts}, and spilled the least amount of food (\textit{Spills}) when using Kiri-Spoon.}
        \label{fig:results}
        \vspace{-3 ex}
    \end{center}
\end{figure*}


So far we have discussed the design and characterization of Kiri-Spoon and validated our proposed deformation model. Our goal in designing Kiri-Spoon was to make it easy for users to acquire and transfer food without spilling it in robot-assisted feeding scenarios. To test this, we conducted a user study comparing the performance of Kiri-Spoon to a traditional spoon and fork in a proof-of-concept feeding task, focusing only on the picking and transfer of food items.




\p{Task and Experimental Setup}
We attached the feeding utensils to the end-effector of a 6-DoF UR5 robot arm mounted on a table. Participants teleoperated the robot and feeding utensils using a hand-held joystick to perform two types of tasks: (i) scooping food from a bowl and (ii) picking food from a plate. For the scooping task, we filled the bowl with foods of varying sizes, including black beans, uncooked pasta, skittles, and almonds. For the picking task, we included foods with different shapes such as cheese cubes, sausage, and grapes. In both tasks, participants had to successfully acquire the food from the bowl or plate and transfer it to another container placed across the table.

\p{Independent Variables}
Participants performed both tasks twice, once using the \textbf{traditional} utensils, spoon and fork, and once using \textbf{Kiri-Spoon} equipped with the food-safe kirigami sheet fabricated in Section~\ref{sec:design}.  The spoon and fork were used to acquire food 
from the \textbf{bowl} and \textbf{plate}, respectively, while Kiri-Spoon was used for both tasks. For scooping, Kiri-Spoon was oriented upwards like a traditional spoon, and for picking, it was oriented downwards like a soft gripper, as shown in Figure~\ref{fig:front}. 

\p{Dependent Variables}
To determine the effectiveness of each feeding utensil, we measured the time taken to acquire the food (\textit{Picking Time}), the changes in the orientation of the feeding utensil (\textit{Orient}), the total weight of food transferred to the other container (\textit{Transferred Weight}), and the number of attempts where the food was spilled (\textit{Spills}). 

To compute \textit{Picking Time} we measured the total duration for which the end-effector remained within $10$ cm of the bowl or plate during each attempt. If the duration exceeded $30$s, we considered it as a \textit{Timeout}. We calculated \textit{Orient} by summing the absolute changes in the pitch of the feeding utensil during acquisition and transfer. Higher values of \textit{Picking Time}, \textit{Timeout}, and \textit{Orient} suggest greater difficulty in controlling the feeding utensil. Whereas, higher values of \textit{Transferred Weight} and lower values of \textit{Spills} indicate that users were more successful in performing the task.

We also administered a 7-point Likert scale survey to assess the subjective experience of users with each feeding utensil in both tasks. Our survey questions were organized into four multi-item scales: how \textit{easy} the utensil was for picking up, how \textit{easy} the utensil was for transferring, 
how much \textit{damage} the utensil caused to the food,
and how much \textit{time} users felt they spent positioning the robot. 
Users also provided their overall preference for either traditional utensils or Kiri-Spoon after all interactions.


\p{Participants and Procedure}
We recruited 12 able-bodied participants (5 female, ages 21.5 $\pm$ 10.5 years) from the Virginia Tech community. All participants provided informed written consent as per the university guidelines (IRB $\# 22$-$308$). 
We followed a within-subject design where each participant performed both the scooping (\textbf{bowl}) and picking (\textbf{plate}) tasks with each type of feeding utensil (\textbf{traditional} and \textbf{Kiri-Spoon}) --- interacting four times with the robot.

Before each interaction, users were given two attempts to practice picking and transferring the food to familiarize themselves with the joystick commands, the motion of the robot arm, and the feeding utensil. During the interaction, users had $5$ attempts to acquire and transfer the food. 
After each interaction, the users answered a survey to report their subjective experience of using the feeding utensil. 
We counterbalanced the order in which users interacted with the feeding utensils, such that half of the participants performed both tasks with the traditional utensils first while the other half started both tasks with Kiri-Spoon. 

\p{Hypotheses}
We hypothesized that:

\begin{adjustwidth}{0.5cm}{0.5cm}
\p{H1} \textit{Users will spill the least amount of food when using Kiri-Spoon.}

\p{H2} \textit{Users will find it easier to control Kiri-Spoon than the traditional utensils.}

\end{adjustwidth}

\p{Results}
The objective results of our user study are summarized in Figure~\ref{fig:results}. We found that participants were able to transfer slightly more food with the Kiri-Spoon ($M=39.54, SD=12.66$) compared to traditional utensils ($M=37.63, SD=20.92$) although the difference was not significant.
On average, users were able to transfer $7.4$ grams more food with Kiri-Spoon than the traditional spoon, but they transferred $3.6$ grams less food with it than the fork. This occurred because users could pick multiple food items more frequently with the fork than with Kiri-Spoon.

While the amount of food transferred was similar for both types of feeding utensils, the food was held more securely by Kiri-Spoon. A paired t-test revealed that users spilled the food in significantly fewer interactions ($t(59)=4.190, p<0.001$) with Kiri-Spoon than with the traditional utensils. This result supports \textbf{H1}. 
Here we observed that all spills for the traditional utensils were when the users were operating the spoon. There were no spills with the traditional fork because the food items chosen in this study were able to firmly adhere to the surface of the fork. All spills for the Kiri-Spoon occurred while scooping the food before it was encapsulated, with no spills during transfer. 


Users had similar \textit{Picking Times} ($t(23)=-1.858$, $p=0.084$) with the traditional utensils ($M=9.44, SD = 3.23$) and Kiri-Spoon ($M=9.37, SD=3.33$). It is important to note that a portion of the picking time for Kiri-Spoon was spent actuating the kirigami structure to secure the food, reducing its time advantage over traditional utensils.
Despite this, there were $5$ \textit{Timeouts} when using the traditional utensils while there were none for Kiri-Spoon.

To compare how much users maneuvered the robot for each feeding utensil, we recorded the total translational and rotational movements during each attempt. While users had similar amounts of translational motion with each feeding utensil, there was a large disparity in their rotational movement (\textit{Orient}). Users had to rotate the robot's end effector significantly more ($t(23)=3.097$, $p=0.005$) when controlling the traditional utensils ($M=6.32$, $SD=4.63$) than Kiri-Spoon ($M=3.58$, $SD=4.63$). 

\begin{figure}[t]
    \begin{center}
        \includegraphics[width=1.0\columnwidth]{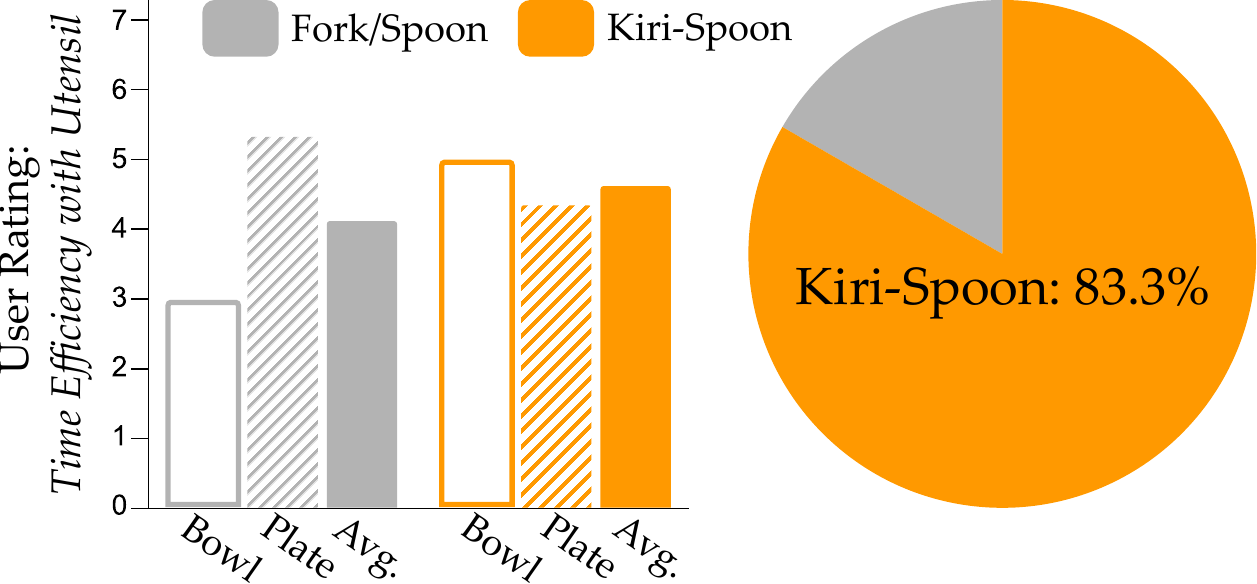}
        \caption{Subjective results from our user study. (Left) Users responded to a 7-point Likert scale survey; the results suggest that users felt like they were more time efficient with Kiri-Spoon than when using traditional utensils. The pie chart displays that, overall, ten of twelve users preferred using the Kiri-Spoon to the traditional utensils across both plate and bowl tasks.}
        \label{fig:subjective}
    \end{center}
\end{figure}

Subjectively, users felt that they spent slightly less time and effort aligning the Kiri-Spoon compared to the traditional utensils. Figure~\ref{fig:subjective} shows the average ratings provided by users for the efficiency of controlling each feeding utensil. These results, combined with objective results for \textit{Picking Time}, \textit{Timeout}, and \textit{Orient} provide partial support for \textbf{H2}.

\p{Discussion} 
Overall, our results demonstrate that Kiri-Spoon is more robust in collecting and transferring food than traditional utensils. Ten out of the twelve participants reported that they preferred using Kiri-Spoon over the spoon and fork. Through open-ended responses in the survey, users mentioned that while "\textit{it was easy to use the fork to pick up}",
"\textit{it was easier overall to use the Kiri-Spoon to grab every kind of food}". Users also stated that the "\textit{biggest difference in efficiency} [was between] \textit{the Kiri-Spoon and the traditional spoon}". Lastly, in addition to these results, we would also like to highlight that Kiri-Spoon is a single device that functions both as a spoon and a fork --- eliminating the need for switching the utensils between tasks.

%% file: 6_conclusion.tex
\section{Conclusion}

In this paper we introduced Kiri-Spoon, a novel utensil for robot-assisted feeding.
Kiri-Spoon is composed of a kirigami sheet that deforms into a $3$D bowl with adjustable curvature: by actuating Kiri-Spoon the robot controls this curvature to rapidly encapsulate and release food items.
We modeled the mechanics and geometry of Kiri-Spoon as a four-bar linkage with springs along the major and minor axes.
Our validation experiments suggest that this model accurately predicts the shape of the kirigami structure and the force needed to actuate the Kiri-Spoon.
Finally, we conducted a study where able-bodied participants controlled a robot arm to pick up and carry diverse food items.
Our results suggest that the ability of Kiri-Spoon to wrap around foods prevented items from spilling, making it easier for humans to orient the robot and transfer foods as compared to standard utensils.

\p{Limitations and Future Work}
As mentioned in Section~\ref{sec:userstudy}, participants often succeeded when using the fork during our user studies. 
The effectiveness of the fork may have been overly inflated by the food items we selected within our experiments; for example, \cite{feng2019robot} report that slippery foods like bananas can easily slide off of forks.
In our future work we want to revisit the comparison between Kiri-Spoon and forks across a wider range of food items.

One fundamental limitation of our current Kiri-Spoon design is that the discrete ribbons create gaps in the spoon surface.
This could allow small food particles to fall from the bottom of the kirigami structure, and would also fail when faced with liquid foods (e.g., soups).
One promising solution here is to coat the top of the kirigami structure with a thin elastic membrane.
This membrane would then deform with the rest of the Kiri-Spoon, maintaining the ability to change curvature while preventing any food items from falling in the gaps between the ribbons.
Our future work will develop this membrane structure.